\documentclass[11pt,a4paper]{article}
\usepackage[hyperref]{naaclhlt2019}
\usepackage{caption}
\usepackage{booktabs}
\usepackage{array, makecell}
\usepackage{multirow}
\usepackage{times}
\usepackage{latexsym}
\usepackage{graphicx}

\usepackage{url}
\aclfinalcopy 
\def\aclpaperid{} 

\usepackage{amsmath,amsfonts,bm}

\def\eqref#1{equation~\ref{#1}}

\def\1{\bm{1}}

\DeclareMathAlphabet{\mathsfit}{\encodingdefault}{\sfdefault}{m}{sl}
\SetMathAlphabet{\mathsfit}{bold}{\encodingdefault}{\sfdefault}{bx}{n}

\newcommand{\fairseq}{\textsc{fairseq}}
\newcommand{\customtilde}{\raise.17ex\hbox{$\scriptstyle\sim$}}

\makeatletter
\renewcommand*{\@fnsymbol}[1]{\ensuremath{\ifcase#1\or * \or \dagger\or \ddagger\or
    \mathsection\or \mathparagraph\or \|\or **\or \dagger\dagger
    \or \ddagger\ddagger \else\@ctrerr\fi}}
\makeatother

\usepackage{siunitx}

\title{\fairseq{}: A Fast, Extensible Toolkit for Sequence Modeling}

\author{
Myle Ott$^\bigtriangleup$\Thanks{\hbox{ }equal contribution}
\quad Sergey Edunov$^{\bigtriangleup*}$
\quad Alexei Baevski$^\bigtriangleup$
\quad Angela Fan$^\bigtriangleup$
\quad Sam Gross$^\bigtriangleup$ \\
\quad {\bf Nathan Ng}$^\bigtriangleup$ 
\quad {\bf David Grangier}$^\bigtriangledown$\Thanks{\hbox{ }Work done while at Facebook AI Research.}
\quad {\bf Michael Auli}$^\bigtriangleup$ \\
\quad $^\bigtriangleup$ Facebook AI Research\\
\quad $^\bigtriangledown$ Google Brain
}

\date{}

\begin{document}
\maketitle
\begin{abstract}
\fairseq{} is an open-source sequence modeling toolkit that allows researchers and developers to train custom models for translation, summarization, language modeling, and other text generation tasks.
The toolkit is based on PyTorch and supports distributed training across multiple GPUs and machines.
We also support fast mixed-precision training and inference on modern GPUs.
A demo video can be found here: \url{https://www.youtube.com/watch?v=OtgDdWtHvto}.
\end{abstract}

\section{Introduction}

Neural sequence-to-sequence models have been successful on a variety of text generation tasks, including machine translation, abstractive document summarization, and language modeling.
Accordingly, both researchers and industry professionals can benefit from a fast and easily extensible sequence modeling toolkit.

There are several toolkits with similar basic functionality, but they differ in focus area and intended audiences.
For example, OpenNMT~\cite{opennmt} is a community-built toolkit written in multiple languages with an emphasis on extensibility.
MarianNMT~\cite{mariannmt} focuses on performance and the backend is written in C++ for fast automatic differentiation.
OpenSeq2Seq~\cite{openseq2seq} provides reference implementations for fast distributed and mixed precision training.
Tensor2tensor~\cite{tensor2tensor} and Sockeye~\cite{sockeye} focus on production-readiness.

In this paper, we present \fairseq{}, a sequence modeling toolkit written in PyTorch that is fast, extensible, and useful for both research and production.
\fairseq{} features:
(i) a common interface across models and tasks that can be extended with user-supplied plug-ins (\textsection\ref{sec:design});
(ii) efficient distributed and mixed precision training, enabling training over datasets with hundreds of millions of sentences on current hardware (\textsection\ref{sec:implementation});
(iii) state-of-the-art implementations and pretrained models for machine translation, summarization, and language modeling (\textsection\ref{sec:use_cases});
and (iv) optimized inference with multiple supported search algorithms, including beam search, diverse beam search~\cite{vijayakumar2016diverse}, and top-k sampling.
\fairseq{} is distributed with a BSD license and is available on GitHub at \url{https://github.com/pytorch/fairseq}.

\section{Design}\label{sec:design}

\paragraph{Extensibility.}
\fairseq{} can be extended through five types of user-supplied plug-ins, which enable experimenting with new ideas while reusing existing components as much as possible.

\paragraph{Models} define the neural network architecture and encapsulate all learnable parameters.
Models extend the \texttt{BaseFairseqModel} class, which in turn extends \texttt{torch.nn.Module}.
Thus any \fairseq{} model can be used as a stand-alone module in other PyTorch code.
Models can additionally predefine named \emph{architectures} with common network configurations (e.g., embedding dimension, number of layers, etc.).
We also abstracted the methods through which the model interacts with the generation algorithm, e.g., beam search, through step-wise prediction. 
This isolates model implementation from the generation algorithm.

\paragraph{Criterions} compute the loss given the model and a batch of data, roughly: \texttt{loss = criterion(model, batch)}.
This formulation makes criterions very expressive, since they have complete access to the model.
For example, a criterion may perform on-the-fly generation to support sequence-level training~\cite{edunov2018classical} or online backtranslation~\cite{edunov2018backtranslation,lample2018phrase}.
Alternatively, in a mixture-of-experts model, a criterion may implement EM-style training and backpropagate only through the expert that produces the lowest loss~\cite{shen2019mixture}.

\paragraph{Tasks} store dictionaries, provide helpers for loading and batching data and define the training loop.
They are intended to be immutable and primarily interface between the various components.
We provide tasks for translation, language modeling, and classification.

\paragraph{Optimizers} update the model parameters based on the gradients.
We provide wrappers around most PyTorch optimizers and an implementation of Adafactor~\cite{shazeer2018adafactor}, which is a memory-efficient variant of Adam.

\paragraph{Learning Rate Schedulers} update the learning rate over the course of training.
We provide several popular schedulers, e.g., the inverse square-root scheduler from~\citet{vaswani2017transformer} and cyclical schedulers based on warm restarts~\cite{loshchilov2016sgdr}.

\paragraph{Reproducibility and forward compatibility.}

\fairseq{} includes features designed to improve reproducibility and forward compatibility.
For example, checkpoints contain the full state of the model, optimizer and dataloader, so that results are reproducible if training is interrupted and resumed.
\fairseq{} also provides forward compatibility, i.e., models trained using old versions of the toolkit will continue to run on the latest version through automatic checkpoint upgrading.

\section{Implementation} \label{sec:implementation}

\fairseq{} is implemented in PyTorch and it provides efficient batching, mixed precision training, multi-GPU as well as multi-machine training.

\paragraph{Batching.}
There are multiple strategies to batch input and output sequence pairs~\citep{morishita2017wmt}.
\fairseq{} minimizes padding within a mini-batch by grouping source and target sequences of similar length. 
The content of each mini-batch stays the same throughout training, however mini-batches themselves are shuffled randomly every epoch.
When training on more than one GPU or machine, then the mini-batches for each worker are likely to differ in the average sentence length which results in more representative updates.

\begin{figure}
\begin{center}
\includegraphics[width=0.9\linewidth]{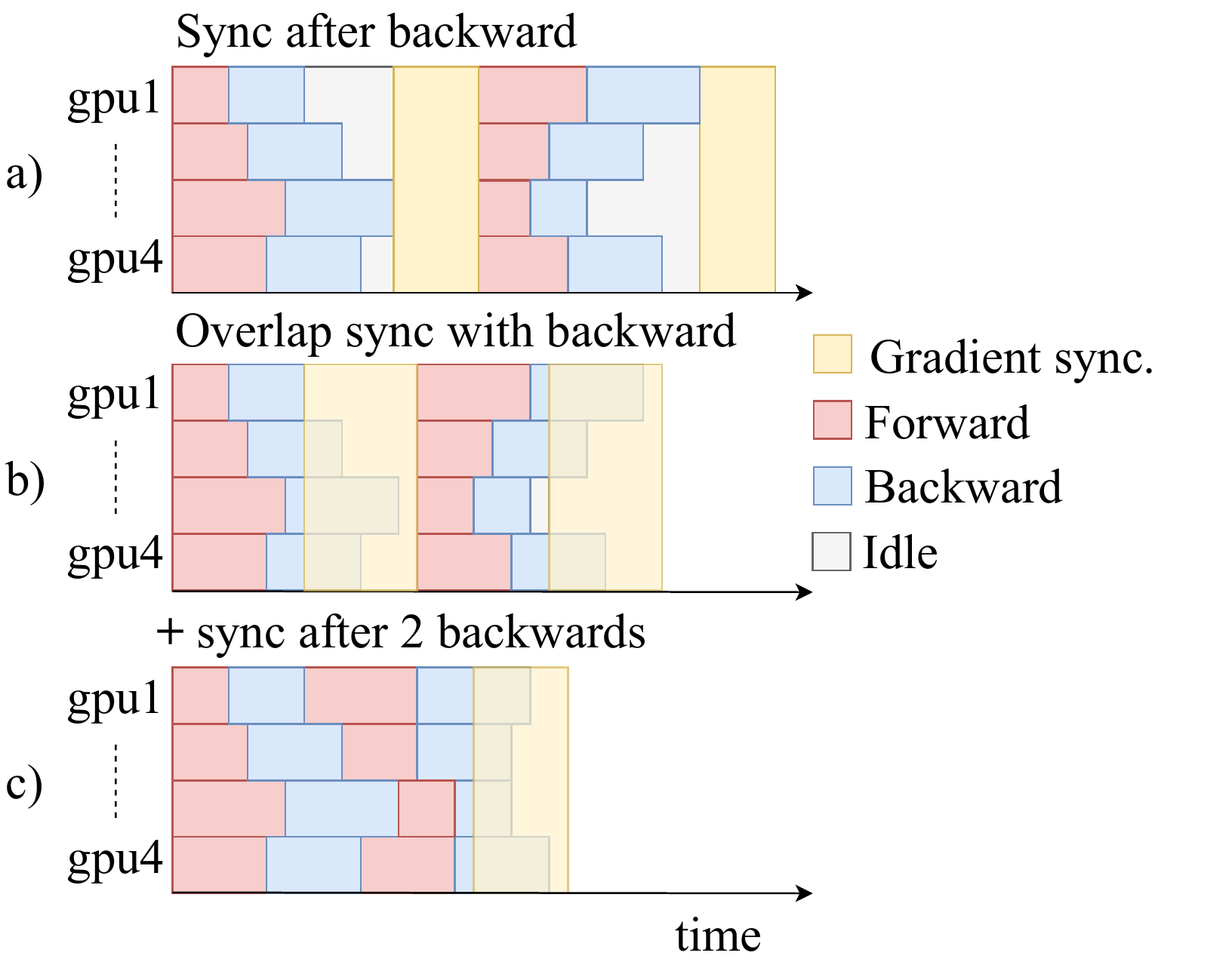}
\end{center}
\caption{Illustration of (a) gradient synchronization and idle time during training, (b) overlapping back-propagation (backward) with gradient synchronization to improve training speed, (c) how accumulating gradient updates can reduce variance in processing time and reduce communication time. 
}
\label{fig:overlap}
\end{figure}

\paragraph{Multi-GPU training.}
\fairseq{} uses the NCCL2 library and \texttt{torch.distributed} for inter-GPU communication. 
Models are trained in a synchronous optimization setup where each GPU has a copy of the model to process a sub-batch of data after which gradients are synchronized between GPUs; all sub-batches constitute a mini-batch.
Even though sub-batches contain a similar number of tokens, we still observe a high variance in processing times.
In multi-GPU or multi-machine setups, this results in idle time for most GPUs while slower workers are finishing their work (Figure~\ref{fig:overlap} (a)).
\fairseq{} mitigates the effect of stragglers by overlapping gradient synchronization between workers with the backward pass and by accumulating gradients over multiple mini-batches for each GPU~\citep{ott2018scaling}.

Overlapping gradient synchronization starts to synchronize gradients of parts of the network when they are computed. 
In particular, when the gradient computation for a layer finishes, \fairseq{} adds the result to a buffer. 
When the size of the buffer reaches a predefined threshold, the gradients are synchronized in a background thread while back-propagation continues as usual (Figure~\ref{fig:overlap} (b)). 
Next, we accumulate gradients for multiple sub-batches on each GPU which reduces the variance in processing time between workers since there is no need to wait for stragglers after each sub-batch (Figure~\ref{fig:overlap} (c)). 
This also increases the effective batch size but we found that models can still be trained effectively~\citep{ott2018scaling}.

\paragraph{Mixed precision.}
Recent GPUs enable efficient half precision floating point (FP16) computation. \fairseq{} provides support for both full precision (FP32) and FP16 at training and inference. 
We perform all forward-backward computations as well as the all-reduce for gradient synchronization between workers in FP16. However, the parameter updates remain in FP32 to preserve accuracy.
\fairseq{} implements dynamic loss scaling~\citep{narang2018iclr} in order to avoid underflows for activations and gradients because of the limited precision offered by FP16.
This scales the loss right after the forward pass to fit into the FP16 range while the backward pass is left unchanged.
After the FP16 gradients are synchronized between workers, we convert them to FP32, restore the original scale, and update the weights.

\paragraph{Inference.}

\begin{table}[t]
\centering
\begin{tabular}{lr}
\toprule
                               & Sentences/sec \\
\midrule
\fairseq{} FP32 & 88.1 \\
\fairseq{} FP16 & \bf{136.0} \\
\bottomrule
\end{tabular}
\caption{Translation speed measured on a V100 GPU on the test set of the standard WMT'14 English-German benchmark using a big Transformer model.}
\label{tab:inference}
\end{table}

\fairseq{} provides fast inference for non-recurrent models \citep{gehring2017convs2s,vaswani2017transformer,fan2018hierarchical,wu2019pay} through incremental decoding, where the model states of previously generated tokens are cached in each active beam and re-used.
This can speed up a na\"ive implementation without caching by up to an order of magnitude, since only new states are computed for each token.
For some models, this requires a component-specific caching implementation, e.g., multi-head attention in the Transformer architecture. 

During inference we build batches with a variable number of examples up to a user-specified number of tokens, similar to training.
\fairseq{} also supports inference in FP16 which increases decoding speed by 54\% compared to FP32 with no loss in accuracy (Table \ref{tab:inference}).

\section{Applications} \label{sec:use_cases}

\fairseq{} has been used in many applications, such as machine translation~\citep{gehring2017convs2s,edunov2018classical,edunov2018backtranslation,chen2018stable,ott:uncertainty:2018,song2018double,wu2019pay}, language modeling~\citep{dauphin2017convlm,baevski2018adp}, abstractive document summarization~\citep{fan2018controllable,liu2018controlling,narayan2018don}, story generation~\citep{fan2018hierarchical,fan2019strategies}, error correction~\citep{chollampatt2018multilayer}, multilingual sentence embeddings~\cite{artetxe2018massively}, and dialogue~\citep{miller2017parlai,dinan2019wizard}.

\subsection{Machine translation}

We provide reference implementations of several popular sequence-to-sequence models which can be used for machine translation, including LSTM~\cite{luong2015effective}, convolutional models~\cite{gehring2017convs2s,wu2019pay} and Transformer~\cite{vaswani2017transformer}.

We evaluate a ``big" Transformer encoder-decoder model on two language pairs, WMT English to German (En--De) and WMT English to French (En--Fr).
For En--De we replicate the setup of~\citet{vaswani2017transformer} which relies on WMT'16 for training with 4.5M sentence pairs, we validate on newstest13 and test on newstest14. The 32K vocabulary is based on a joint source and target byte pair encoding (BPE; \citealt{bpe}).
For En--Fr, we train on WMT'14 and borrow the setup of~\citet{gehring2017convs2s} with 36M training sentence pairs. We use newstest12+13 for validation and newstest14 for test. The 40K vocabulary is based on a joint source and target BPE.

We measure case-sensitive tokenized BLEU with multi-bleu~\citep{multibleu} and de-tokenized BLEU with SacreBLEU\footnote{SacreBLEU hash: \texttt{\scriptsize
\mbox{BLEU+case.mixed+lang.en-\{de,fr\}+}\\
\mbox{numrefs.1+smooth.exp+test.wmt14/full+tok.13a+version.1.2.9}}}~\citep{post2018sacrebleu}.
All results use beam search with a beam width of 4 and length penalty of 0.6, following~\citealt{vaswani2017transformer}. \fairseq{} results are summarized in Table~\ref{tab:testwmt}.
We reported improved BLEU scores over \citet{vaswani2017transformer} by training with a bigger batch size and an increased learning rate~\cite{ott2018scaling}.

\begin{table}[t]
\centering
\begin{tabular}{lr r}
\toprule
                               & En--De   & En--Fr \\
\midrule
a. \citet{gehring2017convs2s}    &  25.2    & 40.5 \\
b. \citet{vaswani2017transformer}&  28.4    & 41.0 \\
c. \citet{ahmed2017WeightedTN}   &  28.9    & 41.4 \\
d. \citet{shaw2018relpos}        &  29.2    & 41.5 \\
\midrule
\fairseq{} Transformer base      & 28.1     & 41.1 \\
\fairseq{} Transformer big       & {\bf 29.3}    & {\bf 43.2} \\
~~~{\it detok. SacreBLEU}   	& {\it 28.6}     & {\it 41.4} \\
{\it 8 GPU training time}   & {\it \customtilde 12 h} & {\it \customtilde 73 h}\\
{\it 128 GPU training time}   & {\it \customtilde 1.3 h} & {\it \customtilde 7.2 h}\\
\bottomrule
\end{tabular}
\caption{BLEU on news2014 for WMT English-German (En--De) and English-French (En--Fr). All results are based on WMT'14 training data, except for En--De (b), (c), (d) and our models which were trained on WMT'16. Train times based on V100 GPUs. }
\label{tab:testwmt}
\end{table}

\subsection{Language modeling}

\begin{table}
\centering
\begin{tabular}{lr}
\toprule
& Perplexity \\ \midrule
\citet{grave2016cache} & 40.8 \\
\citet{dauphin2017convlm}  & 37.2 \\
\citet{merity2018lm} & 33.0 \\
\citet{hebbian} & 29.2 \\
\midrule
\fairseq{} Adaptive inputs & \bf{18.7} \\  
\bottomrule
\end{tabular}
\caption{Test perplexity on WikiText-103 (cf. Table~\ref{tab:gbw_best}).
}
\label{tab:wiki_best}
\end{table}
 
\begin{table}[t]
\centering
\begin{tabular}{lr}
\toprule
& Perplexity \\ \midrule
\citet{dauphin2017convlm} & 31.9 \\
\citet{jozefowicz2016lm} & 30.0 \\
\citet{shazeer2017} & 28.0 \\
\midrule
\fairseq{} Adaptive inputs & \bf 23.0 \\ 
\bottomrule
\end{tabular}
\caption{Test perplexity on the One Billion Word benchmark. Adaptive inputs share parameters with an adaptive softmax.
}
\label{tab:gbw_best}
\end{table}

\fairseq{} supports language modeling with gated convolutional models~\citep{dauphin2017convlm} and Transformer models~\citep{vaswani2017transformer}. 
Models can be trained using a variety of input and output representations, such as standard token embeddings, convolutional character embeddings~\citep{kim2016character}, adaptive softmax \citep{grave2017icml}, and adaptive inputs~\citep{baevski2018adp}.
We also provide tutorials and pre-trained models that replicate the results of~\citet{dauphin2017convlm} and~\citet{baevski2018adp} on WikiText-103 and the One Billion Word datasets. 

We evaluate two Transformer language models, which use only a decoder network and adaptive input embeddings,
following~\citet{baevski2018adp}.
The first model has 16 blocks, inner dimension 4K and embedding dimension 1K; results on WikiText-103 are in Table~\ref{tab:wiki_best}.
The second model has 24 blocks, inner dimension 8K and embedding dimension 1.5K; results on the One Billion Word benchmark are in Table~\ref{tab:gbw_best}.

\subsection{Abstractive document summarization}

Next, we experiment with abstractive document summarization where we use a base Transformer to encode the input document and then generate a summary with a decoder network.
We use the CNN-Dailymail dataset~\citep{hermann2015teaching,nallapati2016abstractive} of news articles paired with multi-sentence summaries. 
We evaluate on the full-text version with no entity anonymization \cite{see2017get}; we truncate articles to 400 tokens~\citep{see2017get}.
We use BPE with 30K operations to form our vocabulary following~\citet{fan2018controllable}. 
To evaluate, we use the standard \textsc{rouge} metric \cite{lin2004rouge} and report \textsc{rouge-1}, \textsc{rouge-2}, and \textsc{rouge-l}. 
To generate summaries, we follow standard practice in tuning the minimum output length and disallow repeating the same trigram \cite{paulus17}. 
Table~\ref{tab:abs} shows results of \fairseq{}. 
We also consider a configuration where we input pre-trained language model representations to the encoder network and this  language model was trained on newscrawl and CNN-Dailymail, totalling 193M sentences.

\begin{table}
\centering
\begin{tabular}{lccc}
\toprule
& \multicolumn{3}{c}{ROUGE}\\
& 1 & 2 & L\\
\toprule
\citet{see2017get} & 39.5 & 17.3 & 36.4 \\
\citet{gehrmann2018bottom} & 41.2 & 18.7 & 38.3 \\
\midrule
\fairseq{} & 40.1 & 17.6 & 36.8 \\
+ pre-trained LM & \bf 41.6 & \bf 18.9 & \bf 38.5 \\
\bottomrule
\end{tabular}
\caption{Abstractive summarization results on the full-text version of CNN-DailyMail dataset.}
\label{tab:abs}
\end{table}

\section{Conclusion}

We presented \fairseq{}, a fast, extensible toolkit for sequence modeling that is scalable and suitable for many applications. 
In the future, we will continue the development of the toolkit to enable further research advances.

\section*{Acknowledgements}

We thank Jonas Gehring for writing the original Lua/Torch version of fairseq.

\bibliography{master}
\bibliographystyle{acl_natbib}

\end{document}